\documentclass[10pt,conference]{IEEEtran}

\usepackage[utf8]{inputenc}
\usepackage[T1]{fontenc}
\usepackage{amsmath,amssymb,amsfonts}
\usepackage{booktabs}
\usepackage{graphicx}
\usepackage{xcolor}
\usepackage{hyperref}
\usepackage{cleveref}
\usepackage{algorithm}
\usepackage{algpseudocode}
\usepackage{tikz}
\usetikzlibrary{arrows,shapes,positioning,shadows,trees}

\usepackage{listings}
\usepackage{upquote}
\usepackage{microtype}

\definecolor{bg}{rgb}{0.98,0.98,0.98}
\definecolor{comment}{rgb}{0.4,0.6,0.4}
\definecolor{keyword}{rgb}{0.0,0.4,0.8}
\definecolor{string}{rgb}{0.6,0.0,0.0}

\hypersetup{
    colorlinks=true,
    linkcolor=blue,
    filecolor=magenta,
    urlcolor=cyan,
    citecolor=blue
}

\title{KL3M Tokenizers: A Family of Domain-Specific and Character-Level Tokenizers for Legal, Financial, and Preprocessing Applications}

\author{
    \IEEEauthorblockN{Michael J. Bommarito II}
    \IEEEauthorblockA{ALEA Institute\thanks{Email: hello@aleainstitute.ai}}
    \IEEEauthorblockA{Stanford CodeX}
    \and
    \IEEEauthorblockN{Daniel Martin Katz}
    \IEEEauthorblockA{Illinois Tech - Chicago Kent Law}
    \IEEEauthorblockA{Bucerius Law School}
    \IEEEauthorblockA{ALEA Institute}
    \IEEEauthorblockA{Stanford CodeX}

    \and
    \IEEEauthorblockN{Jillian Bommarito}
    \IEEEauthorblockA{ALEA Institute}
}

\begin{document}

\maketitle
\begin{abstract}
We present the KL3M tokenizers, a family of specialized tokenizers for legal, financial, and governmental text. Despite established work on tokenization, specialized tokenizers for professional domains remain understudied. Our paper offers two main contributions to this area.

First, we introduce domain-specific BPE tokenizers for legal, financial, and governmental text. Our kl3m-004-128k-cased tokenizer uses 9-17\% fewer tokens than GPT-4o and Llama3 for domain-specific documents, despite having a smaller vocabulary. For specialized terminology, our cased tokenizer is even more efficient, using up to 83\% fewer tokens for legal terms and 39\% fewer tokens for financial terms.

Second, we develop character-level BPE tokenizers (4K, 8K, and 16K vocabulary sizes) for text correction tasks like OCR post-processing. These tokenizers keep consistent token boundaries between error-containing and correct text, making it easier for models to learn correction patterns.

These tokenizers help professional applications by fitting more text in context windows, reducing computational needs, and preserving the meaning of domain-specific terms. Our analysis shows these efficiency gains directly benefit the processing of long legal and financial documents. We release all tokenizers and code through GitHub and Hugging Face to support further research in specialized tokenization.\footnote{For correspondence or assistance accessing our data and models: hello@aleainstitute.ai}
\end{abstract}

\section{Introduction}

Tokenization — the process of converting raw text into discrete tokens — is a fundamental component of modern language models \cite{brown2020language, chowdhery2023palm} that has significant impacts on model performance \cite{rust2020good}. While Byte-Pair Encoding (BPE) \cite{sennrich2016neural, gage1994new} and similar subword tokenization approaches are widely used across domains, recent work has questioned their universal optimality, especially when trained on general corpora \cite{bostrom2020byte}.

In this work, we present the results of our research on tokenizers we developed for the KL3M dataset and models.  This research suggests that specialized domains like law and applications like OCR correction can significantly benefit from custom tokenization approaches.

As an example, consider a legal citation such as \texttt{Fed. R. Civ. P. 56(a)}.  This citation refers to a critical rule in U.S. Federal court procedure that allows a party to ask the court to resolve a case or issue without a full trial.  The presence of this citation is an important indicator for classification tasks, and when drafting, it is critical that motions and briefs properly invoke the rule.  Similarly, financial documents are full of terms and abbreviations that are key to extraction tasks, such \texttt{EBITDA} or \texttt{diluted}.

\begin{table*}[h!]
\centering
\small
\caption{Tokenization comparison across domains}
\label{tab:token-compare}
\begin{tabular}{p{2.5cm}p{7cm}p{7cm}}
\toprule
\textbf{Tokenizer} & \textbf{Legal Text} & \textbf{Financial Text} \\
& \texttt{Fed. R. Civ. P. 56(a)} & \texttt{EBITDA increased by 14.3\%} \\
\midrule
kl3m-004-128k-cased & ["Fed.", " ", "R.", " ", "Civ.", " ", "P.", " 56", "(a)"] & ["EBITDA", " increased", " by", " 14", ".", "3", "\%"] \\
\midrule
kl3m-004-char-8k-cased & ["F", "ed", ".", " R", ".", " C", "iv", ".", " P", ".", " 56", "(", "a", ")"] & ["EB", "IT", "DA", " in", "cre", "as", "ed", " by", " 14", ".", "3", "\%"] \\
\midrule
GPT-4o & ["Fed", ".", " R", ".", " Civ", ".", " P", ".", " 56", "(a)"] & ["EB", "IT", "DA", " increased", " by", " ", "14", ".", "3", "\%"] \\
\midrule
LLaMA 3 & ["Fed", ".", " R", ".", " Civ", ".", " P", ".", " 56", "(a)"] & ["EB", "IT", "DA", " increased", " by", " ", "14", ".", "3", "\%"] \\
\midrule
GPT-2 & ["Fed", ".", " R", ".", " Civ", ".", " P", ".", " 56", "(a)"] & ["E", "BIT", "DA", " increased", " by", " 14", ".", "3", "\%"] \\
\midrule
RoBERTa & ["Fed", ".", " R", ".", " Civ", ".", " P", ".", " 56", "(a)"] & ["E", "BIT", "DA", " increased", " by", " 14", ".", "3", "\%"] \\
\bottomrule
\end{tabular}
\end{table*}

Unfortunately, as highlighted in Table \ref{tab:token-compare}, even the largest tokenizers from frontier labs fail to efficiently capture such language.  For example, both \texttt{cl100k\_base}, the tokenizer behind \texttt{gpt-4o}, and the LLaMA 3 tokenizer require three tokens for \texttt{EBITDA} or \texttt{diluted} each.  In the case of legal citations, these tokenizers also fail to capture the fact that each token is an abbreviation, for example, by splitting the letter R from the abbreviating period.

While these differences may seem minor, anyone who has reviewed embedding layers or investigated inference pathologies like hallucination will appreciate the impact that such tokenizer issues can cause.  Furthermore, tokenizer efficiency is a critical factor in the amount of text that can fit into a model's context window.  While this has implications for the cost of training and inference generally, it is especially important for the legal and financial domain, where documents often have both more words and longer words than in other contexts.

As part of our research on datasets and models in the legal domain, we investigated a number of alternative approaches to tokenization that might address issues like the examples above.  This research began with the \texttt{kl3m-001-32k} tokenizer and then branched into two separate groups of models: domain-specific BPE tokenizers and character-level BPE tokenizers.  

Our domain-specific KL3M tokenizers (\texttt{kl3m-003-64k}, \texttt{kl3m-004-128k-cased}, \texttt{kl3m-004-128k-uncased}) are 9-17\% more efficient than \texttt{cl100k\_base}, the \texttt{gpt-4o} tokenizer, despite having a substantially smaller vocabulary. The cased variant in particular (\texttt{kl3m-004-128k-cased}) provides excellent performance across the legal and financial domains while maintaining case sensitivity, which is critical for many domain tasks.

Our character-level BPE tokenizers (\texttt{kl3m-004-char-4k}, \texttt{kl3m-004-char-8k}, \texttt{kl3m-004-char-16k}), though less thoroughly researched, have been instrumental in training our OCR correction models, such as the 500M parameter \texttt{kl3m-004-correction-001} model.

The dual approach of domain-specific and character-level tokenizers within the KL3M family addresses complementary needs we faced in the KL3M project: efficient representation for the most common tasks and character-level precision for error correction in pretrain and RAG applications. Although our work focuses on legal, financial, and governmental domains, we believe similar approaches could potentially be relevant for other specialized fields. 

All KL3M tokenizers are available on GitHub (\url{https://github.com/alea-institute/kl3m-tokenizers} and Hugging Face (\url{https://huggingface.co/alea-institute}), along with the source code, training data, and related models.  The source code for this paper, including \LaTeX and replication for figures and tables, is available at \url{https://github.com/alea-institute/kl3m-tokenizer-paper/}.
\section{Background and Related Work}

\subsection{Tokenization Evolution}

Tokenization approaches in NLP have evolved from simple word-splitting \cite{webster1992tokenization} to sophisticated algorithms optimized for specific linguistic properties and computational requirements.\cite{mielke2021between}

Word-level tokenization, common in early NLP systems \cite{webster1992tokenization}, faced significant challenges with out-of-vocabulary words and morphological variance. Character-level approaches \cite{ma2x020charbert} addressed vocabulary limitations but generated longer sequences and lost semantic coherence. The tokenization-free CANINE approach \cite{clark2022canine} eliminates the need for explicit tokenization but required specialized model architectures.

Modern NLP systems predominantly using subword tokenization, with Byte-Pair Encoding (BPE) \cite{sennrich2016neural,gage1994new} became the de facto standard for large language models \cite{brown2020language}. Alternative methods include SentencePiece \cite{kudo2018sentencepiece}, which offer language-agnostic tokenization such as WordPiece used in BERT \cite{devlin2019bert}. These approaches balance vocabulary size constraints with morphological awareness.

\subsection{Byte-Pair Encoding}

Byte-Pair Encoding (BPE) was originally developed as a data compression algorithm \cite{gage1994new} and later adapted for NLP tokenization \cite{sennrich2016neural}. The algorithm can be summarized as:

\begin{enumerate}
    \item Start with a vocabulary of individual characters;
    \item Identify the most frequent adjacent character pair and merge them into a new token;
    \item Repeat this process iteratively until a desired vocabulary size or alternative stopping condition is reached.
\end{enumerate}

BPE offers several key benefits, including its ability to effectively manage rare words and morphological variations by breaking them down into smaller, meaningful subword units.  This subword tokenization approach also allows BPE to maintain a fixed vocabulary size, which is crucial for computational efficiency in NLP models.  Furthermore, BPE is designed to preserve frequent words as single tokens, while less common words are segmented into subwords, striking a balance between efficient representation and the ability to handle out-of-vocabulary terms.  

However, standard BPE approaches have limitations. Bostrom and Durrett \cite{bostrom2020byte} demonstrated that BPE can be suboptimal for language model pretraining due to its frequency-based merging, which may not align with the linguistic intuitions of typical users. 

\subsection{Domain-Specific Tokenization}

While general-purpose tokenizers are designed for broad applicability, specialized domains can benefit significantly from tailored tokenization approaches.

In the biomedical domain, BioBERT \cite{lee2020biobert} demonstrated that even with standard tokenization, domain-specific pretraining improves performance. Similarly, SciBERT \cite{beltagy2019scibert} for scientific text and BERTweet \cite{nguyen2020bertweet} for social media text show that domain adaptation at the model level improves performance, but they do not specifically address tokenization challenges.

Legal and financial domains present unique tokenization challenges due to specialized terminology, citation formats, and document structures. Chalkidis et al. \cite{chalkidis2020legal} introduced Legal-BERT, which adapted BERT for legal text but retained the original WordPiece tokenizer, focusing adaptation on the model parameters rather than the tokenization approach.

In the financial domain, Araci \cite{araci2019finbert} proposed FinBERT for financial sentiment analysis, while Mansar et al. \cite{mansar2021finsim} organized the FinSim shared task focusing on financial terminology. However, these works primarily address model adaptation rather than tokenization optimization.

To date, we are not aware of any significant effort to build and evaluate domain-specific tokenizers for legal, regulatory and financial texts.  

\subsection{Character-Level and Other Approaches}

Character-level models and tokenizers have seen renewed interest for specialized applications. Ma et al. \cite{ma2020charbert} proposed CharBERT, which incorporates character-level information into the model architecture. Clark et al. \cite{clark2022canine} introduced CANINE, a tokenization-free encoder that operates directly on Unicode characters.

These approaches are particularly relevant for tasks requiring character-level understanding, such as spelling correction, OCR post-processing, and handling of noisy text. However, they typically require specific model architectures designed to work with character-level input, rather than providing character-aware tokenizers for existing architectures.

\subsection{Tokenization Evaluation}

Evaluating tokenizer performance presents unique challenges. Rust et al. \cite{rust2020good} assessed how tokenizer quality impacts model performance for multilingual applications. The authors found that the choice of tokenizer significantly affects downstream performance, particularly for languages with complex morphology.

Most tokenizer evaluations focus on intrinsic measures like vocabulary coverage or extrinsic measures of downstream task performance. There is limited work on evaluating tokenizers specifically for domain-specific applications, with most evaluations focusing on general text or cross-lingual scenarios.

Our work builds on these foundations but differs in several important ways. Unlike previous approaches that adapt models but not tokenizers to domains, we directly address the tokenization challenges in legal and financial text. 
\section{Methodology}

This section details our approach to designing and training the KL3M tokenizers, including the data sources and tokenizer design.

\subsection{Data Sources}

The KL3M tokenizers were trained on a diverse corpus of legal, financial, and governmental documents from our KL3M dataset. A fundamental principle of our approach was to use only data that is free from copyright or licensing restrictions, ensuring that the training data and resulting tokenizers can be used without restriction.

Primary data sources included:

\begin{itemize}
    \item US government documents and websites produced by the executive or legislative branches under 17 USC 105
    \item EU government documents produced under CC-BY or Decision 2011/833
    \item US state and federal court opinions and associated documents
    \item Publicy-traded and registered company filings, including financial reports and legal agreements
    \item Granted patents filed with the USPTO
\end{itemize}

These datasets can be browsed under our Hugging Face account: \url{https://huggingface.co/alea-institute}.

\subsection{Tokenizer Design}

While our goal is to address a number of issues with traditional BPE tokenizers, we also wanted to ensure that our tokenizers could be easily used by ourselves and others.  Therefore, we constrained our implementations to be compatible with the \texttt{tokenizers} BPE implementation, also available through the \texttt{transformers} library, to maximize compatibility with existing libraries and pipelines.

\subsubsection{Original Tokenizer: kl3m-001-32k}

Our first tokenizer, \texttt{kl3m-001-32k}, was designed as a test bed for our first models, \texttt{kl3m-002-170m} and \texttt{kl3m-003-1.7b}.  In addition to its domain-specific training corpus, this tokenizer also featured a number of alterations:

\begin{itemize}
  \item no space (Ġ) prefixing
  \item a small set of custom tokens, including limited whitespace, Markdown, HTML, JSON, XML, and numeric tokens
  \item special tokens for both MLM and CLM tasks
  \item power-of-2 padding
\end{itemize}

While we successfully trained models up to 1.7B on this tokenizer, our experience, especially with the decreased efficiency from the removal of space prefixing and struggles with OCR correction, led us to split our research into two families of tokenizers - domain-specific BPEs and character-level BPEs.

\subsubsection{Domain-Specific BPE}
While \texttt{kl3m-001-32k}'s small vocabulary had advantages for memory usage, we unsurprisingly found that 32K tokens was inefficient for many typical generative use cases.  Furthermore, we found that custom tokens were extremely useful and increased reliability in a number of important use cases.  As a result, we substantially increased the size of our vocabulary to 64K and 128K, increased the size and breadth of custom tokens, standardized on NFKC normalization, and introduced an uncased variant of our 128K tokenizer for embedding models.

\subsubsection{Character Tokenizers}
Conversely, for error correction and normalization tasks like OCR post-processing, we found that the 32K vocabularly was likely too large, requiring substantially more parameters to learn basic operations like  character confusion or transposition.  Given that we needed to use these models at the scale of pretrain corpora, model size and speed was an extremely important consideration.

To address this, we developed specialized character-level tokenizers with 4K, 8K, and 16K vocabulary sizes.  These tokenizers rely on a modified training technique that constrains the maximum merged token length to emphasize character-level patterns.  The 4K and 8K tokenizers have a maximum token length of three characters, while the 16K tokenizer allows up to four characters per token.

Character-level tokenizers are particularly valuable for text error correction in legal and financial documents, where errors can significantly alter meaning. Errors in these domains frequently occur in predictable patterns from multiple sources including OCR, manual transcription, and user entry:

\begin{itemize}
    \item Character confusions: similar-looking characters (e.g., "c"/"e", "5"/"S", "0"/"o", "l"/"1")
    \item Spacing errors: inappropriate spaces or joined words (e.g., "S tates" instead of "States")
    \item Character transpositions: reversed character order (e.g., "Teh" instead of "The")
    \item Domain-specific substitutions: legal/financial symbols (e.g., "§"/"S", "¶"/"P")
    \item Typographical errors: common keyboard-based mistakes (e.g., adjacent key hits, double letters)
    \item Phonetic errors: spelling based on pronunciation (e.g., "eksept" instead of "except")
\end{itemize}

Drawing on character-aware approaches like CharBERT \cite{ma2020charbert} and CANINE \cite{clark2022canine}, but with important modifications, our character tokenizers are optimized for different use cases:

\begin{itemize}
    \item \textbf{kl3m-004-char-4k-cased:} Optimized for pure character-level models and fine-grained spelling correction, similar to character-based approaches in \cite{ma2020charbert}. This tokenizer provides granular character-by-character tokenization ideal for learning exact substitutions (e.g., "c" → "e") in text errors.
    
    \item \textbf{kl3m-004-char-8k-cased:} Balanced approach for general text error correction, with slightly larger character groupings that efficiently handle both character-level errors and common error patterns in various document types.
    
    \item \textbf{kl3m-004-char-16k-cased:} Incorporates domain-specific character sequences for specialized correction tasks while maintaining character-level precision, suitable for more nuanced, domain-specific correction in legal and financial documents.
\end{itemize}

Unlike standard BPE tokenizers that treat text errors as unknown tokens or fragment them inconsistently, our character tokenizers maintain stable token boundaries between incorrect and correct forms. This consistency creates a more direct mapping for transformer models to learn correction patterns, aligning with findings from Wang et al. \cite{wang2022deepstructure} on structure-preserving tokenization.

\subsection{Custom Tokens}

A key aspect of the KL3M tokenizers is the deliberate inclusion of domain-specific and format-specific tokens that might not emerge naturally from BPE training on a general corpus. By explicitly adding these custom tokens, we can steer both tokenizer training and downstream models more easily. 

These custom tokens are grouped into the following categories:

\begin{itemize}
    \item \textbf{Whitespace}: Combinations and repetitions of spaces, tabs, newlines, and carriage returns
    \item \textbf{Markdown}: Common Markdown elements, especially related to document headings, formatting, and lists
    \item \textbf{HTML}: Common HTML tags, including opening and closing tag substrings and attributes
    \item \textbf{JSON}: Common JSON tokens
    \item \textbf{XML}: Common XML tokens
    \item \textbf{Years}: Years from 1776 to 2050
    \item \textbf{Numbers}: Numbers from 1-999
    \item \textbf{Enumerations}: Enumerations such as Roman numerals, including in parenthetical format (e.g., \texttt{(iv)})
    \item \textbf{Citations}: Common legal citations and related tokens derived from the Free Law Project's \texttt{reporters-db}
\end{itemize}

While the 32K tokenizer included some of these custom token groups, the 64K and 128K tokenizers contain many more custom tokens.

\subsection{Power-of-2 Padding}

All KL3M tokenizers are padded to align with powers of 2 during the final stage of training.  In the event that the standard BPE training algorithm stopped before hitting the target vocabulary size, additional whitespace combination tokens (e.g., repeating spaces or newlines) are added until the vocabulary is an exact power.  This provides enhanced efficiency opportunities for storage, computation, and search.

%
\section{Evaluation Methodology}
\label{sec:eval}

This section describes our methodology for evaluating the performance of KL3M tokenizers compared to established tokenizers like \texttt{gpt-4o}, \texttt{LLaMA3}, and \texttt{gpt-2}. We assess three key dimensions: tokenization efficiency, domain term representation, and token size distribution.

\subsection{Datasets and Tokenizers}

We evaluated performance across five diverse datasets selected to represent different domains:

\begin{itemize}
    \item \textbf{US Code}: Federal statutes with specialized legal terminology and structure
    \item \textbf{Congressional Hearings}: Formal political dialogue and legislative terminology
    \item \textbf{Court Documents}: Judicial language with complex citation patterns
    \item \textbf{SEC Filings}: Financial disclosure documents with business and accounting terminology
    \item \textbf{General Content}: Non-specialized texts for baseline comparison
\end{itemize}

Each dataset contains 20-100 documents, providing a representative sample of each domain's linguistic characteristics. Documents were preprocessed to remove markup while preserving text structure.

We compared the following tokenizers:
\begin{itemize}
    \item \textbf{KL3M Standard}: \texttt{kl3m-004-128k-cased}, \texttt{kl3m-004-128k-uncased}, kl3m-003-64k
    \item \textbf{KL3M Character}: kl3m-004-char-4k-cased, kl3m-004-char-8k-cased, kl3m-004-char-16k-cased
    \item \textbf{Comparison}: \texttt{gpt-4o}, \texttt{LLaMA3}, \texttt{gpt-2}
\end{itemize}

\subsection{Evaluation Metrics}

\subsubsection{Tokenization Efficiency}

We measure efficiency using the tokens per character (TPC) ratio across datasets:

\begin{equation}
\text{TPC} = \frac{\text{Number of tokens}}{\text{Number of characters}}
\end{equation}

TPC represents the inverse of compression ratio, with lower values indicating higher efficiency as fewer tokens are needed to represent the same text. This directly impacts computational requirements for processing text.

\subsubsection{Domain Term Representation}

We evaluated how effectively each tokenizer represents domain-specific terminology by measuring token counts for common legal and financial terms. This analysis reveals how well tokenizers capture specialized language patterns, which affects model performance when processing domain-specific content.

\subsubsection{Token Size Distribution}

We analyzed the distribution of token lengths (in characters) across each tokenizer's vocabulary, categorizing tokens as short (1-2 characters), medium (3-6 characters), or long (7+ characters). This distribution provides insights into tokenization strategies and their implications for efficiency and semantic preservation.

\subsubsection{Character Tokenizer Evaluation}

For the specialized KL3M character tokenizers, we conducted additional analyses comparing their tokenization patterns on text containing errors typical of OCR and manual transcription processes. This evaluation demonstrates how consistent character-level tokenization affects error correction capabilities.
\section{Results}
\label{sec:results}

This section presents our evaluation results comparing KL3M tokenizers with \texttt{gpt-4o}, \texttt{LLaMA3}, and \texttt{gpt-2} tokenizers across two key dimensions: tokenization efficiency and domain-specific terminology representation. A detailed analysis of token size distribution is available in Appendix~\ref{app:token_size}.

\subsection{Tokenization Efficiency}

Tokenization efficiency directly impacts both computational resource requirements and context window utilization for transformer models. We measure efficiency using tokens per character (TPC), which represents the inverse of compression ratio—lower values indicate better efficiency as fewer tokens are needed to encode the same text. Table~\ref{tab:token-efficiency} presents a comprehensive comparison across datasets:

\begin{table*}[ht]
\centering
\caption{Token efficiency (tokens per character) across datasets\\\small \textit{Note: Lower values indicate more efficient tokenization (fewer tokens per character).}}
\label{tab:token-efficiency}
\small
\begin{tabular}{lrrrrrr}
\toprule
Dataset & kl3m\mbox{-}004\mbox{-}128k\mbox{-}cased & kl3m\mbox{-}004\mbox{-}128k\mbox{-}uncased & kl3m\mbox{-}003\mbox{-}64k & gpt\mbox{-}4o & llama3 & gpt2 \\
\midrule
Congressional Hearings & 0.2292 & \textbf{0.2231} & 0.2808 & 0.2482 & 0.2475 & 0.3765 \\
Court Documents & 0.2741 & \textbf{0.2692} & 0.2907 & 0.2971 & 0.2986 & 0.3524 \\
General Content & 0.2057 & \textbf{0.2033} & 0.2223 & \textbf{0.2033} & 0.2066 & 0.2116 \\
SEC Filings & 0.1816 & \textbf{0.1804} & 0.1915 & 0.1976 & 0.1992 & 0.3720 \\
US Code & 0.3181 & \textbf{0.3171} & 0.3296 & 0.3716 & 0.3717 & 0.3595 \\
\midrule
Average & 0.2417 & \textbf{0.2386} & 0.2630 & 0.2636 & 0.2647 & 0.3344 \\
\bottomrule
\end{tabular}
\end{table*}

As demonstrated in Table~\ref{tab:token-efficiency}, while \texttt{kl3m-004-128k-uncased} achieves the lowest TPC values across all datasets, our primary \texttt{kl3m-004-128k-cased} model shows nearly equivalent efficiency with particularly significant gains in domain-specific content:

\begin{itemize}
    \item \textbf{Overall efficiency}: Our cased model delivers 9\% average improvement over \texttt{gpt-4o} and \texttt{LLaMA3} (0.2417 vs. 0.2636/0.2647)
    \item \textbf{US Code}: 17\% improvement over \texttt{gpt-4o} and \texttt{LLaMA3} (0.3181 vs. 0.3716/0.3717)
    \item \textbf{Congressional Hearings}: 8\% improvement (0.2292 vs. 0.2482/0.2475)
    \item \textbf{SEC Filings}: 9\% improvement (0.1816 vs. 0.1976/0.1992)
\end{itemize}

These efficiency gains directly translate to practical benefits: for a typical 100K-character legal document, \texttt{kl3m-004-128k-cased} would require approximately 24,170 tokens compared to 26,360 tokens for \texttt{gpt-4o}—a difference that can significantly impact context window utilization. Appendix~\ref{app:token_count} provides the absolute token counts for each tokenizer across all tested datasets, showing that \texttt{kl3m-004-128k-cased} requires 8-10\% fewer tokens than \texttt{gpt-4o} and \texttt{llama3} for the same content. More detailed visualizations of these efficiency metrics can be found in Appendix~\ref{app:token_efficiency_charts}.

Figure~\ref{fig:vocab_size_vs_efficiency} reveals that KL3M tokenizers achieve superior efficiency despite having vocabulary sizes comparable to other tokenizers. This indicates that vocabulary composition and token distribution—not just vocabulary size—are critical factors for tokenization efficiency.

\begin{figure*}[htbp]
    \centering
    \includegraphics[width=0.85\textwidth]{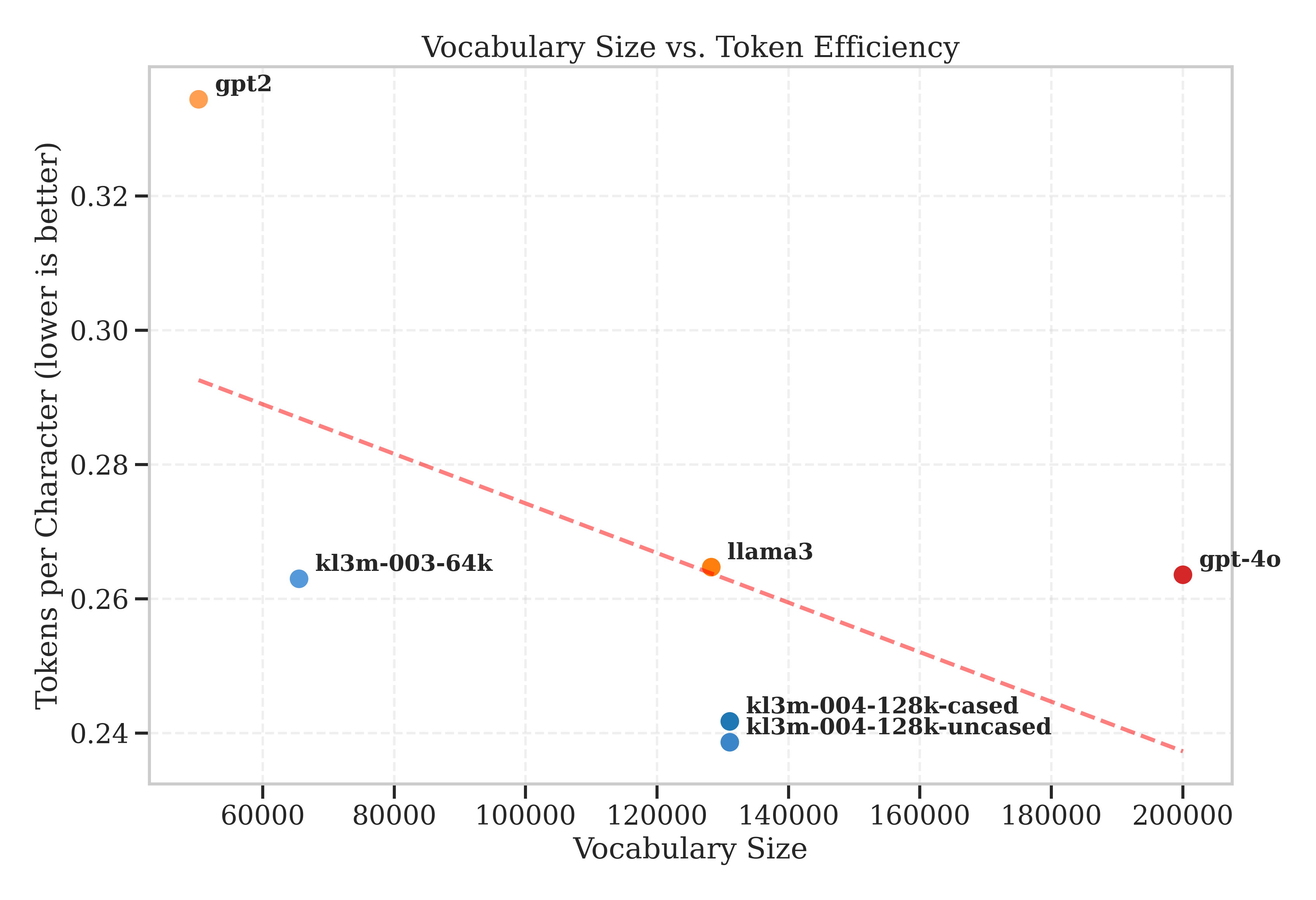}
    \caption{Relationship between vocabulary size and tokenization efficiency. KL3M tokenizers achieve superior efficiency despite having comparable vocabulary sizes to other tokenizers.}
    \label{fig:vocab_size_vs_efficiency}
\end{figure*}

\subsection{Domain Term Representation}

Efficient representation of domain-specific terminology directly impacts how accurately and efficiently models can process specialized content. A key advantage of domain-optimized tokenizers is their ability to represent technical terms with fewer tokens, preserving semantic integrity. Tables~\ref{tab:domain-term-comparison} and \ref{tab:domain-term-aggregate} demonstrate how KL3M tokenizers achieve substantial improvements in this critical dimension:

\begin{table*}[ht]
\centering
\caption{Token count comparison for domain-specific terminology across tokenizers}
\label{tab:domain-term-comparison}
\small
\begin{tabular}{@{}llrrrrrr@{}}
\toprule
Domain & Term & \rotatebox{90}{kl3m\mbox{-}004\mbox{-}128k\mbox{-}cased} & \rotatebox{90}{kl3m\mbox{-}004\mbox{-}128k\mbox{-}uncased} & \rotatebox{90}{gpt\mbox{-}4o} & \rotatebox{90}{llama3} & \rotatebox{90}{roberta\mbox{-}base} & \rotatebox{90}{gpt2} \\
\midrule
Legal & 11 U.S.C. \S 362(a) & \textbf{6} & \textbf{6} & 10 & 11 & 15 & 13 \\
 & res judicata & \textbf{2} & \textbf{2} & 3 & 5 & 6 & 4 \\
 & stare decisis & \textbf{3} & \textbf{3} & 4 & 5 & 7 & 5 \\
 & habeas corpus & \textbf{2} & \textbf{2} & 4 & 5 & 5 & 3 \\
 & certiorari & \textbf{1} & \textbf{1} & 3 & 4 & 5 & 3 \\
 & de novo review & \textbf{3} & \textbf{3} & \textbf{3} & 4 & 6 & 4 \\
 & 28 C.F.R. \S 14.2(a) & \textbf{8} & \textbf{8} & 12 & 13 & 16 & 14 \\
 & 42 U.S.C. \S 1983 & \textbf{5} & \textbf{5} & 9 & 10 & 11 & 9 \\
 & Fed. R. Civ. P. 12(b)(6) & \textbf{10} & \textbf{10} & 14 & 15 & 16 & 14 \\
 & prima facie & \textbf{2} & \textbf{2} & 3 & 5 & 6 & 4 \\
\addlinespace[0.5em]
Financial & EBITDA & \textbf{1} & \textbf{1} & 3 & 4 & 5 & 3 \\
 & P/E ratio & 4 & 4 & \textbf{3} & 4 & 6 & 4 \\
 & 10-K filing & 4 & 4 & \textbf{3} & 4 & 6 & 4 \\
 & SEC Form 8-K & \textbf{5} & \textbf{5} & \textbf{5} & 6 & 7 & \textbf{5} \\
 & quarterly dividend & \textbf{2} & \textbf{2} & 3 & 4 & 5 & 3 \\
 & year-over-year growth & 6 & 6 & \textbf{4} & 5 & 8 & 6 \\
 & Basel III compliance & \textbf{3} & \textbf{3} & 4 & 5 & 6 & 4 \\
 & GAAP accounting & \textbf{2} & \textbf{2} & 3 & 4 & 5 & 3 \\
 & ROI analysis & \textbf{2} & \textbf{2} & \textbf{2} & 3 & 5 & 3 \\
 & market capitalization & \textbf{2} & \textbf{2} & \textbf{2} & 4 & 5 & 3 \\
\bottomrule
\end{tabular}
\end{table*}

\begin{table*}[ht]
\centering
\caption{Average token count by domain across tokenizers}
\label{tab:domain-term-aggregate}
\begin{tabular}{lrrr}
\toprule
Tokenizer & Legal Terms & Financial Terms & Overall \\
\midrule
kl3m\mbox{-}004\mbox{-}128k\mbox{-}cased & \textbf{4.20} & \textbf{3.10} & \textbf{3.65} \\
kl3m\mbox{-}004\mbox{-}128k\mbox{-}uncased & \textbf{4.20} & \textbf{3.10} & \textbf{3.65} \\
gpt\mbox{-}4o & 6.50 & 3.20 & 4.85 \\
llama3 & 7.70 & 4.30 & 6.00 \\
roberta\mbox{-}base & 9.30 & 5.80 & 7.55 \\
gpt2 & 7.30 & 3.80 & 5.55 \\
\bottomrule
\end{tabular}
\end{table*}

The data reveals significant differences in how tokenizers handle specialized terminology:

\begin{itemize}
    \item \textbf{Legal terminology}: \texttt{kl3m-004-128k-cased} encodes terms like "certiorari" and "habeas corpus" in just 1-2 tokens, while \texttt{gpt-4o} and \texttt{LLaMA3} require 3-5 tokens. For complex legal citations (e.g., "42 U.S.C. § 1983"), our cased tokenizer requires 5 tokens compared to 9-10 tokens for other tokenizers.
    
    \item \textbf{Financial terminology}: Terms like "EBITDA" are represented as a single token in \texttt{kl3m-004-128k-cased} but require 3-5 tokens in other tokenizers. Even for complex terms like "Basel III compliance," our cased tokenizer achieves a 25-40\% reduction in token count.
    
    \item \textbf{Overall efficiency}: As shown in Table~\ref{tab:domain-term-aggregate}, \texttt{kl3m-004-128k-cased} requires an average of just 3.65 tokens for domain-specific terms, compared to 4.85 for \texttt{gpt-4o} (33\% more), 6.00 for \texttt{LLaMA3} (64\% more), and 5.55 for \texttt{gpt-2} (52\% more).
\end{itemize}

This more efficient representation offers two key advantages: (1) it reduces computational overhead by requiring fewer tokens for domain-specific content, and (2) it preserves semantic integrity by keeping domain concepts as atomic units rather than fragmenting them into semantically meaningless subwords. This preservation of semantic boundaries is particularly valuable for fine-tuning models on domain-specific tasks where conceptual precision is critical.

\subsection{Character Tokenizer Performance}

Our specialized KL3M character tokenizers (4K, 8K, and 16K variants) represent a distinct approach optimized for text normalization and error correction. Unlike standard tokenizers, these are designed with deliberately constrained vocabulary sizes and maximum token lengths (2-5 characters), resulting in more consistent tokenization patterns that facilitate character-level transformations.

Comparative analysis across these variants revealed that:
\begin{itemize}
    \item The 4K variant provides maximum granularity for pure character-level operations
    \item The 8K variant offers an optimal balance between granularity and efficiency for general text correction
    \item The 16K variant incorporates larger domain-specific character sequences beneficial for specialized correction tasks
\end{itemize}

This approach is particularly valuable for legal and financial documents where errors from OCR, transcription, or manual entry can significantly alter meaning and where preserving exact formatting is critical.

\subsubsection{Text Error Correction Example}

Table \ref{tab:error_example} demonstrates how the character-level tokenizers process text errors compared to standard tokenizers. Consider the text "Thc Vnited S tates 5enate is nesp0nslbe for the" (an error-containing version of "The United States Senate is responsible for the," representative of OCR mistakes, transcription errors, or user typos). 

\begin{table*}[htbp]
    \centering
    \small
    \caption{Character-level tokenization of text errors from test data}
    \label{tab:error_example}
    \begin{tabular}{p{3.2cm}p{5.5cm}p{5.5cm}p{2.5cm}}
        \toprule
        \textbf{Tokenizer} & \textbf{Error Text} & \textbf{Correct Text} & \textbf{Notes} \\
        & \textit{Thc Vnited S tates 5enate is nesp0nslbe for the} & \textit{The United States Senate is responsible for the} & \\
        \midrule
        kl3m-004-char-4k-cased & 
        ["Th", "c", " V", "n", "it", "ed", " S", " t", "at", "es", " 5", "en", "ate", 
        " is", " n", "esp", "0", "ns", "l", "be", " f", "or", " t", "he"] & 
        ["The", " Un", "it", "ed", " St", "at", "es", " S", "en", "ate", 
        " is", " re", "sp", "ons", "ib", "le", " f", "or", " t", "he"] & 
        Preserves character positioning \\
        \midrule
        kl3m-004-char-8k-cased & 
        ["Th", "c", " V", "nit", "ed", " S", " t", "at", "es", " 5", "en", "ate", 
        " is", " n", "esp", "0", "ns", "l", "be", " f", "or", " t", "he"] & 
        ["The", " Un", "it", "ed", " St", "at", "es", " S", "en", "ate", 
        " is", " re", "sp", "ons", "ib", "le", " f", "or", " t", "he"] & 
        Balances character groups \\
        \midrule
        kl3m-004-char-16k-cased & 
        ["Th", "c", " V", "n", "ited", " S", " t", "ates", " 5", "en", "ate", 
        " is", " n", "esp", "0", "ns", "l", "be", " for", " the"] & 
        ["The", " Un", "ited", " St", "ates", " Sen", "ate", 
        " is", " re", "sp", "ons", "ible", " for", " the"] & 
        Larger character groupings \\
        \midrule
        gpt-4o & 
        ["Th", "c", " V", "n", "ited", " S", " t", "ates", " ", "5", "en", "ate", 
        " is", " n", "esp", "0", "n", "sl", "be", " for", " the"] & 
        ["The", " Un", "ited", " States", " Sen", "ate", 
        " is", " respons", "ible", " for", " the"] & 
        Less consistent boundaries \\
        \bottomrule
    \end{tabular}
\end{table*}

As shown in Table \ref{tab:error_example}, the character-level tokenizers precisely tokenize each character or small character group in the error text. This fine-grained tokenization allows models to:

\begin{itemize}
    \item Make direct character-to-character mappings (e.g., recognizing "Thc" → "The")
    \item Detect and correct transposition errors (e.g., "nesp0nslbe" → "responsible")
    \item Handle insertion/deletion errors (e.g., "S tates" → "States")
    \item Correct character confusions and substitutions (e.g., "5enate" → "Senate", "0" → "o")
\end{itemize}

As Table \ref{tab:error_example} illustrates, the key advantage of KL3M character tokenizers compared to standard BPE approaches is their consistent tokenization patterns between error-containing and correct text forms. While standard tokenizers like \texttt{gpt-4o} produce radically different token boundaries when handling errors, our character tokenizers maintain stable segmentation patterns that enable more effective learning of correction mappings.

This structured approach to character-level tokenization benefits several practical use cases:

\begin{itemize}
    \item \textbf{OCR post-processing}: Correcting scanning errors in legal documents and financial statements where errors can significantly alter meaning
    \item \textbf{Transcription normalization}: Standardizing court transcripts, congressional records, and regulatory filings where formatting consistency is critical
    \item \textbf{Document digitization}: Converting legacy documents where character-level errors are common but systematic
    \item \textbf{Legal citation standardization}: Ensuring consistent formatting of case citations, statutory references, and regulatory citations
\end{itemize}

The efficiency gains from this approach are particularly valuable in large-scale document processing workflows where both accuracy and computational efficiency are critical considerations.
\section{Discussion}

Our evaluation results demonstrate the advantages of domain-specific tokenization for legal, financial, and governmental text. This section discusses the practical implications of our findings and examines key limitations and challenges of specialized tokenizers.

\subsection{Practical Benefits for Professional Applications}

The efficiency advantages of KL3M tokenizers translate to several practical benefits:

\begin{itemize}
    \item \textbf{Expanded effective context window:} When processing legal or financial documents with thousands of domain-specific terms and citations, the 9-17\% overall efficiency improvement and up to 83\% improvement for specific terminology substantially expands the effective context window. This allows models to process more complete documents without truncation.
    
    \item \textbf{Reduced computational costs:} The reduced token count directly translates to lower computational requirements for training and inference. For applications processing millions of legal or financial documents, this efficiency can yield significant computational savings.
    
    \item \textbf{Enhanced cross-document reasoning:} By representing citations and references more efficiently and coherently, KL3M-004-128k-cased enables better tracking of references across documents, which is crucial for legal research, financial analysis, and regulatory compliance tasks.
\end{itemize}

For fine-tuning applications, our domain-specific tokenizers also deliver notable benefits. When fine-tuning general models on legal or financial texts, tokenization mismatches between pre-training and fine-tuning data can disrupt performance. By using tokenizers specifically designed for professional content during both pre-training and fine-tuning, this mismatch can be reduced or eliminated, leading to better specialized model performance.

\subsection{Limitations and Challenges}

\subsubsection{Domain Specificity vs. Generality}

A core limitation of specialized tokenizers lies in the inherent trade-off between domain specificity and broader applicability. By allocating vocabulary space to domain-specific terms, these tokenizers leave less room for general terms, which can enhance performance on targeted content but may compromise effectiveness on more general text. While our analysis showed that KL3M-004-128k-cased performed well on general content, more extensive evaluation across diverse general text types would be needed to fully assess this trade-off.

Furthermore, KL3M tokenizers, designed specifically for legal and financial domains, may prove less efficient when applied to other specialized fields such as medicine or chemistry, limiting their cross-domain utility. Additionally, like any fixed-vocabulary tokenizer, they may struggle to efficiently handle novel terminology that emerges after the training data is set, potentially reducing their adaptability to evolving language.

\subsubsection{Implementation Challenges}

Several technical challenges surfaced during implementation. One key issue was striking the right balance between cased and uncased variants; although we provide both options, the best choice depends heavily on the specific application requirements. Another difficulty arose in custom token selection, where identifying domain-specific tokens required considerable expert knowledge, suggesting that more systematic methods could enhance the process. 

Furthermore, while we obtained high-quality, copyright-free legal and financial texts for training, acquisition of additional content may be a significant limitation in subsequent efforts to further improve tokenizer quality and performance. The quality and representativeness of training data remain crucial factors in tokenizer performance.

\subsubsection{Ecosystem Compatibility}

Although our implementation ensures compatibility with the Hugging Face ecosystem, broader compatibility issues exist. Certain model architectures rely on assumptions about tokenization that may not align with the design of specialized tokenizers, posing integration challenges. Additionally, using our tokenizers with existing pre-trained models requires careful alignment or adjustments to embedding layers to ensure proper functionality.

The NLP ecosystem tends to prioritize support for a handful of widely used tokenizers, which could restrict the availability of tools and libraries compatible with our specialized approach. Despite these challenges, the significant efficiency advantages demonstrated in our evaluation suggest that overcoming these implementation and compatibility issues would be worthwhile for professional applications working with legal and financial documents.

\subsection{Future Work}

Several promising directions for future work emerge from our research:

\begin{itemize}
    \item \textbf{Downstream task evaluation:} Providing public and reproducible experiments on downstream training tasks like masked language modeling (MLM) and causal language modeling (CLM) would quantify the impact of domain-specific tokenization on model quality, with particular emphasis on the OCR/error correction capabilities of our character-level tokenizers.
    
    \item \textbf{Tokenizer swapping:} Investigating the impact and methodology of tokenizer swapping for established pretrained models like \texttt{LLaMA3} could enable existing models to benefit from domain-specific tokenization without complete retraining, potentially offering an efficient path to domain adaptation.
    
    \item \textbf{Custom token extensions:} Further refinement of the custom token selection process could enhance domain coverage, particularly through development of systematic methods to identify high-value domain-specific tokens that maximize efficiency gains across diverse professional documents.
    
    \item \textbf{Non-English professional language support:} Extending our approach to common non-English professional languages (e.g., EU legal frameworks in French and German, international financial reporting terminology) would address the growing need for multilingual domain-specific NLP capabilities in global regulatory and business contexts.
\end{itemize}
\section{Conclusion}

In this paper, we presented the KL3M family of tokenizers for legal, financial, and governmental text processing. Our research demonstrates significant advantages of domain-specific tokenization for professional applications, with five key contributions:

\begin{enumerate}
    \item A methodology for domain-specific tokenization that balances learned tokens with curated domain-specific additions, building on foundational BPE work \cite{sennrich2016neural}.
    
    \item A dual-approach tokenizer family with standard BPE variants (64K-128K vocabulary) for efficient representation and character-level variants (4K-16K vocabulary) for OCR correction and text normalization.
    
    \item Quantitative efficiency improvements: Our analysis shows that \texttt{kl3m-004-128k-cased} achieves excellent tokenization efficiency with an average of 0.2417 tokens per character across datasets, representing a 9\% improvement over \texttt{gpt-4o} (0.2636) and \texttt{LLaMA3} (0.2647) despite being 35\% smaller than \texttt{gpt-4o}. While our uncased variant offers slightly better efficiency (0.2386 tokens per character), the cased model maintains important case distinctions necessary for many professional applications.
    
    \item Evidence that domain-specific tokenization preserves semantic integrity of specialized terminology and citation patterns critical to professional understanding, with \texttt{kl3m-004-128k-cased} requiring an average of just 3.65 tokens for domain-specific terms compared to 4.85 for \texttt{gpt-4o} (33\% more tokens) and 6.00 for \texttt{LLaMA3} (64\% more tokens).
    
    \item A rigorous evaluation framework for tokenizer performance in specialized domains, extending previous evaluation approaches \cite{rust2020good}.
\end{enumerate}

The efficiency advantages of \texttt{kl3m-004-128k-cased} are particularly pronounced for specialized content. For US Code, our cased tokenizer achieved 0.3181 tokens per character compared to 0.3716 for \texttt{gpt-4o} (a 17\% improvement) and 0.3717 for \texttt{LLaMA3} (a 17\% improvement). For SEC filings, our cased model achieved 0.1816 tokens per character compared to 0.1976 for \texttt{gpt-4o} (a 9\% improvement) and 0.1992 for \texttt{LLaMA3} (a 10\% improvement). This demonstrates that even against state-of-the-art tokenizers like \texttt{gpt-4o} and \texttt{LLaMA3}, domain-specialized tokenizers can provide substantial efficiency gains.

Our domain term analysis provides even more compelling evidence for the value of specialized tokenization. For legal terminology, \texttt{kl3m-004-128k-cased} required an average of just 4.20 tokens per term, while \texttt{gpt-4o} needed 6.50 tokens (55\% more) and \texttt{LLaMA3} required 7.70 tokens (83\% more). For financial terms, the advantage persisted with KL3M requiring 3.10 tokens compared to 3.20 for \texttt{gpt-4o} and 4.30 for \texttt{LLaMA3} (39\% more). Key legal terms like "certiorari" required just 1 token in KL3M compared to 3 tokens in \texttt{gpt-4o} and 4 tokens in \texttt{LLaMA3}, while "11 U.S.C. § 362(a)" required 6 tokens in KL3M versus 10 tokens in \texttt{gpt-4o} and 11 tokens in \texttt{LLaMA3}.

These efficiency gains directly translate to context window utilization, computational efficiency, and lower inference costs. For large legal or financial documents with thousands of domain-specific terms and citations, the 9-17\% overall efficiency improvement can substantially expand the effective context window, allowing more complete document understanding without truncation. This advantage becomes particularly important for long-document reasoning and question answering, where parsing and understanding references across a lengthy document is essential.

Domain-specific tokenization represents an important frontier in NLP research. Building on domain-specific model work like Legal-BERT \cite{chalkidis2020legal} and FinBERT \cite{araci2019finbert}, our results suggest that tokenization customization can provide substantial benefits complementary to model specialization. Future work could extend to multilingual legal systems, additional professional domains, and dynamic vocabulary adaptation approaches.

All KL3M tokenizers are publicly available through the Hugging Face Hub under the \texttt{alea-institute/} account to support researchers and practitioners working with legal, financial, and governmental text. The complete source code, including tokenizer training scripts and evaluation tools, is available on GitHub at \url{https://github.com/alea-institute/kl3m-tokenizers}. This open-source release enables full reproducibility and further extension of our work by the research community. Our work establishes that tokenization—far from being a solved problem—remains critical for domain-specific optimization as language models transform professional workflows in specialized domains.

\section{Acknowledgements}
We drafted and revised this paper with the assistance of large language models. All errors or omissions are our own.

\bibliographystyle{IEEEtran}
\bibliography{bibliography/references}

\begin{thebibliography}{10}
\providecommand{\url}[1]{#1}
\csname url@samestyle\endcsname
\providecommand{\newblock}{\relax}
\providecommand{\bibinfo}[2]{#2}
\providecommand{\BIBentrySTDinterwordspacing}{\spaceskip=0pt\relax}
\providecommand{\BIBentryALTinterwordstretchfactor}{4}
\providecommand{\BIBentryALTinterwordspacing}{\spaceskip=\fontdimen2\font plus
\BIBentryALTinterwordstretchfactor\fontdimen3\font minus
  \fontdimen4\font\relax}
\providecommand{\BIBforeignlanguage}[2]{{%
\expandafter\ifx\csname l@#1\endcsname\relax
\typeout{** WARNING: IEEEtran.bst: No hyphenation pattern has been}%
\typeout{** loaded for the language `#1'. Using the pattern for}%
\typeout{** the default language instead.}%
\else
\language=\csname l@#1\endcsname
\fi
#2}}
\providecommand{\BIBdecl}{\relax}
\BIBdecl

\bibitem{brown2020language}
T.~Brown, B.~Mann, N.~Ryder, M.~Subbiah, J.~D. Kaplan, P.~Dhariwal,
  A.~Neelakantan, P.~Shyam, G.~Sastry, A.~Askell \emph{et~al.}, ``Language
  models are few-shot learners,'' \emph{Advances in neural information
  processing systems}, vol.~33, pp. 1877--1901, 2020.

\bibitem{chowdhery2023palm}
A.~Chowdhery, S.~Narang, J.~Devlin, M.~Bosma, G.~Mishra, A.~Roberts, P.~Barham,
  H.~W. Chung, C.~Sutton, S.~Gehrmann \emph{et~al.}, ``Palm: Scaling language
  modeling with pathways,'' \emph{Journal of Machine Learning Research},
  vol.~24, no. 240, pp. 1--113, 2023.

\bibitem{rust2020good}
P.~Rust, J.~Pfeiffer, I.~Vulić, S.~Ruder, and I.~Gurevych, ``How good is your
  tokenizer? on the monolingual performance of multilingual language models,''
  in \emph{Proceedings of the 59th Annual Meeting of the Association for
  Computational Linguistics and the 11th International Joint Conference on
  Natural Language Processing (Volume 1: Long Papers)}, 2021, pp. 3118--3135.

\bibitem{sennrich2016neural}
R.~Sennrich, B.~Haddow, and A.~Birch, ``Neural machine translation of rare
  words with subword units,'' in \emph{Proceedings of the 54th Annual Meeting
  of the Association for Computational Linguistics (Volume 1: Long Papers)},
  2016, pp. 1715--1725.

\bibitem{gage1994new}
P.~Gage, ``A new algorithm for data compression,'' \emph{The C Users Journal},
  vol.~12, no.~2, pp. 23--38, 1994.

\bibitem{bostrom2020byte}
K.~Bostrom and G.~Durrett, ``Byte pair encoding is suboptimal for language
  model pretraining,'' in \emph{Findings of the Association for Computational
  Linguistics: EMNLP 2020}, 2020, pp. 4617--4624.

\bibitem{webster1992tokenization}
J.~J. Webster and C.~Kit, ``Tokenization as the initial phase in nlp,'' in
  \emph{COLING 1992 volume 4: The 14th international conference on
  computational linguistics}, 1992.

\bibitem{mielke2021between}
S.~J. Mielke, Z.~Alyafeai, E.~Salesky, C.~Raffel, M.~Dey, M.~Gall{\'e},
  A.~Raja, C.~Si, W.~Y. Lee, B.~Sagot \emph{et~al.}, ``Between words and
  characters: A brief history of open-vocabulary modeling and tokenization in
  nlp,'' \emph{arXiv preprint arXiv:2112.10508}, 2021.

\bibitem{clark2022canine}
J.~H. Clark, D.~Garrette, I.~Turc, and J.~Wieting, ``Canine: Pre-training an
  efficient tokenization-free encoder for language representation,''
  \emph{Transactions of the Association for Computational Linguistics},
  vol.~10, pp. 73--91, 2022.

\bibitem{kudo2018sentencepiece}
T.~Kudo and J.~Richardson, ``Sentencepiece: A simple and language independent
  subword tokenizer and detokenizer for neural text processing,'' in
  \emph{Proceedings of the 2018 Conference on Empirical Methods in Natural
  Language Processing: System Demonstrations}, 2018, pp. 66--71.

\bibitem{devlin2019bert}
J.~Devlin, M.-W. Chang, K.~Lee, and K.~Toutanova, ``Bert: Pre-training of deep
  bidirectional transformers for language understanding,'' in \emph{Proceedings
  of the 2019 conference of the North American chapter of the association for
  computational linguistics: human language technologies, volume 1 (long and
  short papers)}, 2019, pp. 4171--4186.

\bibitem{lee2020biobert}
J.~Lee, W.~Yoon, S.~Kim, D.~Kim, S.~Kim, C.~H. So, and J.~Kang, ``Biobert: a
  pre-trained biomedical language representation model for biomedical text
  mining,'' \emph{Bioinformatics}, vol.~36, no.~4, pp. 1234--1240, 2020.

\bibitem{beltagy2019scibert}
I.~Beltagy, K.~Lo, and A.~Cohan, ``Scibert: A pretrained language model for
  scientific text,'' in \emph{Proceedings of the 2019 Conference on Empirical
  Methods in Natural Language Processing and the 9th International Joint
  Conference on Natural Language Processing (EMNLP-IJCNLP)}, 2019, pp.
  3615--3620.

\bibitem{nguyen2020bertweet}
D.~Q. Nguyen, T.~Vu, and A.~Tuan~Nguyen, ``Bertweet: A pre-trained language
  model for english tweets,'' in \emph{Proceedings of the 2020 Conference on
  Empirical Methods in Natural Language Processing: System Demonstrations},
  2020, pp. 9--14.

\bibitem{chalkidis2020legal}
I.~Chalkidis, M.~Fergadiotis, P.~Malakasiotis, N.~Aletras, and
  I.~Androutsopoulos, ``Legal-bert: The muppets straight out of law school,''
  in \emph{Findings of the Association for Computational Linguistics: EMNLP
  2020}, 2020, pp. 2898--2904.

\bibitem{araci2019finbert}
D.~Araci, ``Finbert: Financial sentiment analysis with pre-trained language
  models,'' \emph{arXiv preprint arXiv:1908.10063}, 2019.

\bibitem{mansar2021finsim}
Y.~Mansar, J.~Kang, and I.~E. Maarouf, ``The finsim-2 2021 shared task:
  Learning semantic similarities for the financial domain,'' in \emph{Companion
  Proceedings of the Web Conference 2021}, 2021, pp. 288--292.

\bibitem{ma2020charbert}
W.~Ma, Y.~Cui, C.~Si, T.~Liu, S.~Wang, and G.~Hu, ``Charbert: Character-aware
  pre-trained language model,'' in \emph{Proceedings of the 28th International
  Conference on Computational Linguistics}, 2020, pp. 39--50.

\bibitem{wang2022deepstructure}
C.~Wang, X.~Liu, Z.~Chen, H.~Hong, J.~Tang, and D.~Song, ``Deepstruct:
  Pretraining of language models for structure prediction,'' in \emph{Findings
  of the Association for Computational Linguistics: ACL 2022}, 2022, pp.
  803--823.

\end{thebibliography}

\onecolumn

\appendix
\label{app:detailed_results}

\section{Detailed Results}
\label{app:detailed_tables}

\subsection{Token Size Distribution}
\label{app:token_size}

\begin{figure*}[htbp]
    \centering
    \includegraphics[width=0.95\textwidth]{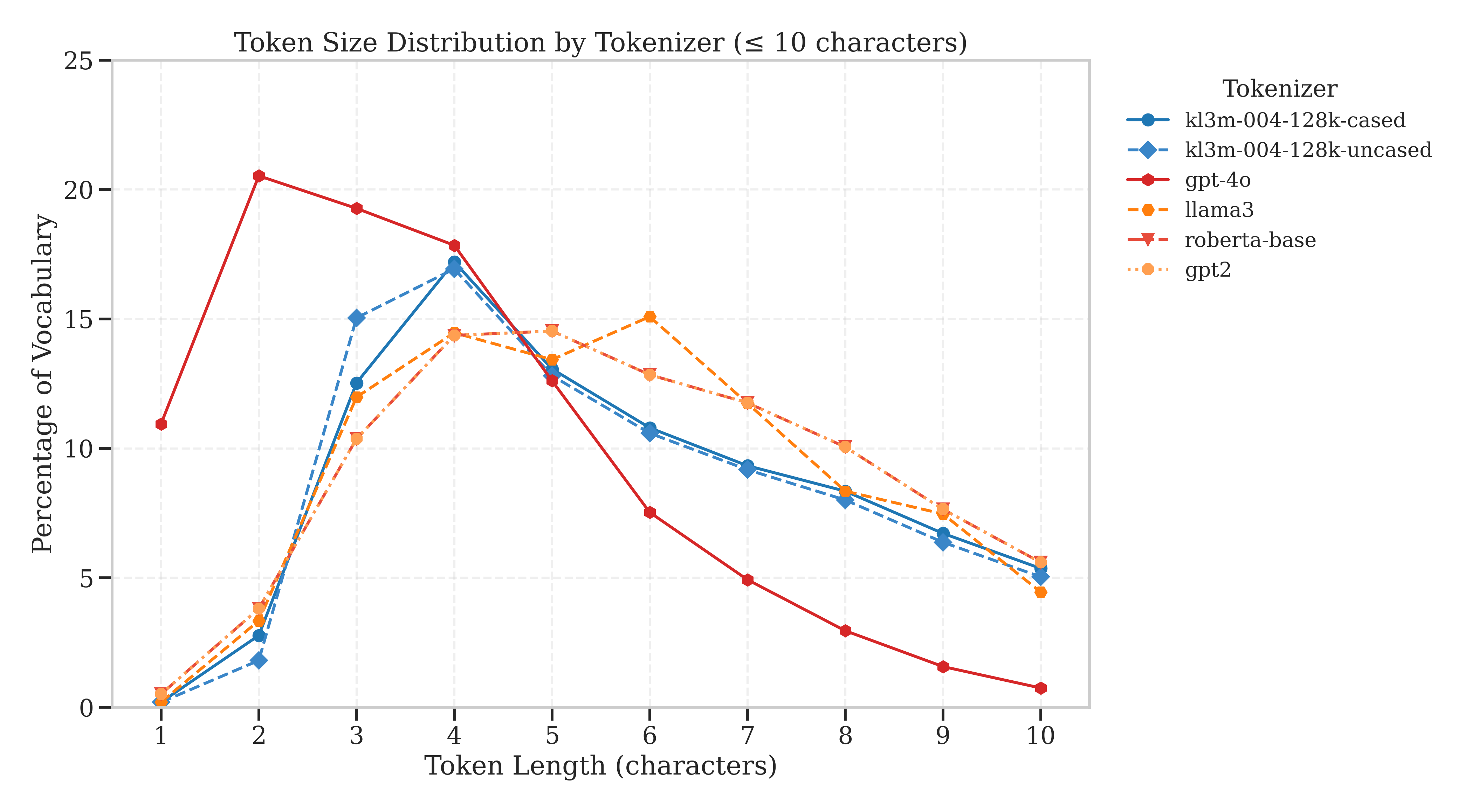}
    \caption{Percentage of vocabulary by token length across tokenizers. KL3M tokenizers show higher percentages of medium-length tokens (3-6 characters) compared to other tokenizers.}
    \label{fig:token_size_comparison}
\end{figure*}

The distribution of token lengths provides important insights into tokenizer design and potential efficiency. Figure~\ref{fig:token_size_comparison} presents a line plot comparing the percentage of vocabulary by token length across tokenizers.

\begin{table*}[htbp]
\centering
\caption{Token size distribution (percentage of vocabulary by character length). Note: For tiktoken models like \texttt{gpt-4o}, only a subset of tokens can be individually decoded, so statistics are based on a partial sample.}
\label{tab:token-size-distribution}
\small
\begin{tabular}{lrrrrrr}
\toprule
Length & kl3m\mbox{-}004\mbox{-}128k\mbox{-}cased & kl3m\mbox{-}004\mbox{-}128k\mbox{-}uncased & gpt\mbox{-}4o & llama3 & roberta\mbox{-}base & gpt2 \\
\midrule
1 & 0.2\% & 0.2\% & 10.9\% & 0.2\% & 0.5\% & 0.5\% \\
2 & 2.8\% & 1.8\% & 20.5\% & 3.3\% & 3.8\% & 3.8\% \\
3 & 12.5\% & 15.0\% & 19.3\% & 12.0\% & 10.4\% & 10.4\% \\
4 & 17.2\% & 16.9\% & 17.8\% & 14.5\% & 14.4\% & 14.4\% \\
5 & 13.1\% & 12.8\% & 12.6\% & 13.4\% & 14.5\% & 14.5\% \\
6 & 10.8\% & 10.6\% & 7.5\% & 15.1\% & 12.8\% & 12.8\% \\
7 & 9.3\% & 9.2\% & 4.9\% & 11.7\% & 11.8\% & 11.8\% \\
8 & 8.3\% & 8.0\% & 3.0\% & 8.4\% & 10.1\% & 10.1\% \\
9 & 6.7\% & 6.4\% & 1.6\% & 7.5\% & 7.6\% & 7.7\% \\
10 & 5.4\% & 5.0\% & 0.7\% & 4.5\% & 5.6\% & 5.6\% \\
\midrule
Total $\leq 5$ & 45.8\% & 46.8\% & 81.2\% & 43.4\% & 43.6\% & 43.6\% \\
Total 6-10 & 40.6\% & 39.2\% & 17.7\% & 47.1\% & 47.9\% & 47.9\% \\
Total $\leq 10$ & 86.3\% & 86.0\% & 99.0\% & 90.5\% & 91.5\% & 91.5\% \\
\bottomrule
\end{tabular}
\end{table*}

KL3M tokenizers, particularly the kl3m-004-128k variants, have a more balanced distribution with a higher percentage of medium-length tokens (3-6 characters). In contrast, \texttt{gpt-4o} has a significantly higher percentage of short tokens (1-2 characters), with over 31

\subsection{Tokenization Efficiency Visualizations}
\label{app:token_efficiency_charts}

\begin{figure*}[htbp]
    \centering
    \includegraphics[width=0.85\textwidth]{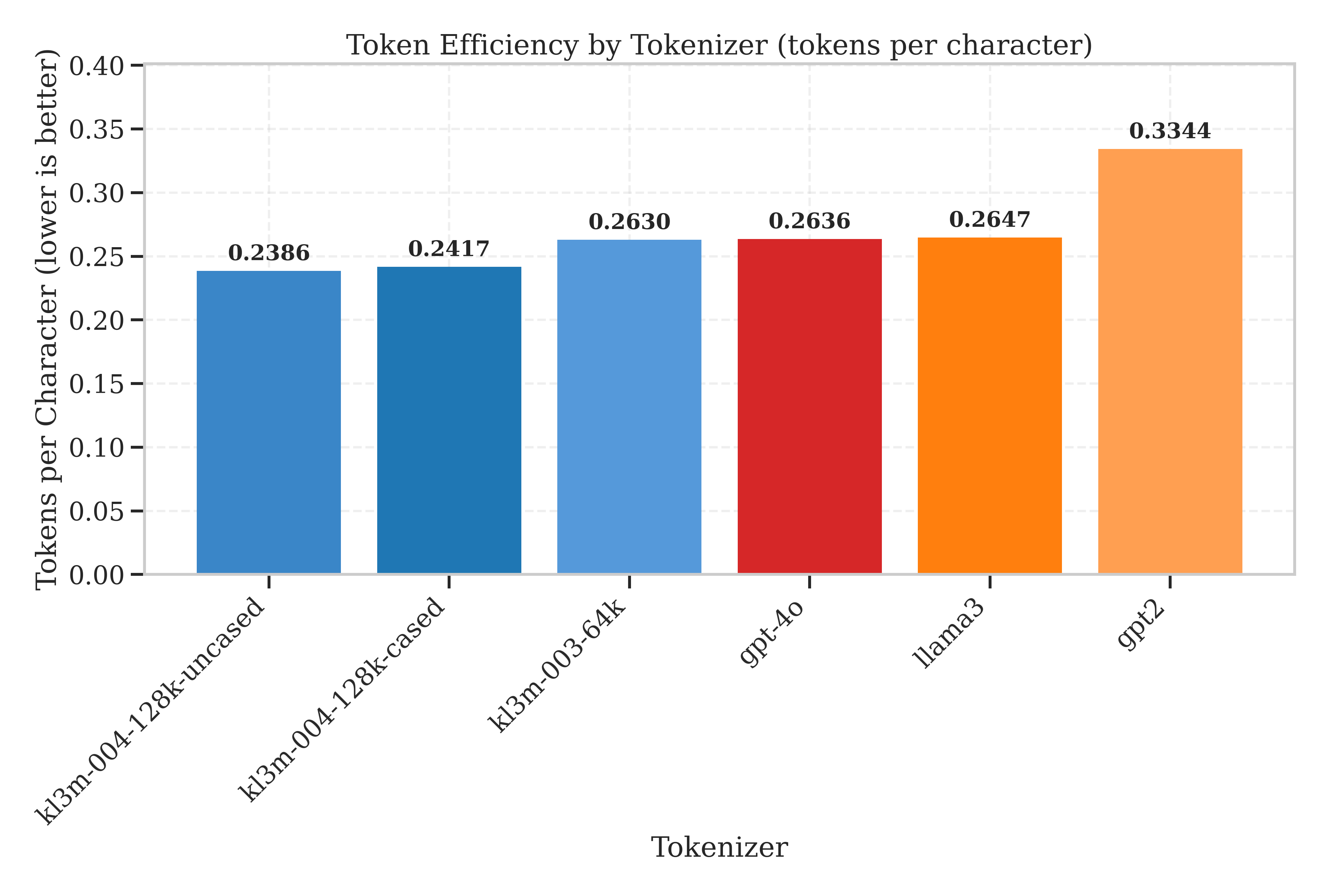}
    \caption{Tokenization efficiency (tokens per character) across datasets. Lower values indicate higher efficiency. KL3M tokenizers consistently demonstrate higher efficiency, particularly for domain-specific content like US Code and Congressional Hearings.}
    \label{fig:token_efficiency_bars}
\end{figure*}

Tokenization efficiency, measured as tokens per character (TPC), represents the inverse of compression ratio. Lower TPC values indicate better compression and higher efficiency, as fewer tokens are needed to encode the same amount of text. Figure~\ref{fig:token_efficiency_bars} provides a visual comparison of tokenization efficiency across datasets for different tokenizers.

\subsection{Total Token Count Comparison}
\label{app:token_count}

Table~\ref{tab:token-count} presents the absolute token counts for each tokenizer across the different datasets, providing a concrete measure of the efficiency improvements offered by the KL3M tokenizers.

\begin{table}[ht]
\centering
\caption{Total token count across datasets\\\small \textit{Note: Lower values indicate more efficient tokenization (fewer tokens to represent the same text).}}
\label{tab:token-count}
\small
\begin{tabular}{lrrrrrr}
\toprule
Dataset & kl3m\mbox{-}004\mbox{-}128k\mbox{-}cased & kl3m\mbox{-}004\mbox{-}128k\mbox{-}uncased & kl3m\mbox{-}003\mbox{-}64k & gpt\mbox{-}4o & llama3 & gpt2 \\
\midrule
Congressional Hearings & 308,169 & \textbf{299,954} & 377,458 & 333,702 & 332,791 & 506,115 \\
Court Documents & 23,682 & \textbf{23,260} & 25,116 & 25,674 & 25,798 & 30,454 \\
General Content & 22,424 & \textbf{22,164} & 24,237 & 22,167 & 22,522 & 23,068 \\
SEC Filings & 243,580 & \textbf{242,057} & 256,913 & 265,078 & 267,267 & 499,126 \\
US Code & 45,099 & \textbf{44,957} & 46,732 & 52,684 & 52,689 & 50,962 \\
\midrule
Total & 642,954 & \textbf{632,392} & 730,456 & 699,305 & 701,067 & 1,109,725 \\
\bottomrule
\end{tabular}
\end{table}

\end{document}